\documentclass{llncs}
\usepackage{graphicx}
\usepackage{float}
\usepackage{cite}
\usepackage{multirow}
\usepackage{url}
\usepackage[utf8]{inputenc} 
\usepackage{amsmath}
\usepackage{amsfonts}
\usepackage[usenames,dvipsnames]{color}
\usepackage{tabularx}

\usepackage{verbatim}

\begin{document}

\title{Learning Word Embeddings from\\the Portuguese Twitter Stream:\\A Study of some Practical Aspects}
\author{Pedro Saleiro$^{1,2}$, Luís Sarmento, Eduarda Mendes Rodrigues$^{1}$,\\Carlos Soares$^{1,3}$, Eugénio Oliveira$^{1,2}$ }
\institute{$^1$FEUP, $^2$LIACC, $^3$INESC TEC, Universidade do Porto, Portugal\\
  \email{pssc@fe.up.pt}
}\maketitle

\begin{abstract}
This paper describes a preliminary study for producing and distributing a large-scale database of embeddings from the Portuguese Twitter stream. We start by experimenting with a relatively small sample and focusing on three challenges: volume of training data, vocabulary size and intrinsic evaluation metrics. Using a single GPU, we were able to scale up vocabulary size from 2048 words embedded and 500K training examples to 32768 words over 10M training examples while keeping a stable validation loss and approximately linear trend on training time per epoch. We also observed that using less than 50\% of the available training examples for each vocabulary size might result in overfitting. Results on intrinsic evaluation show promising performance for a vocabulary size of 32768 words. Nevertheless, intrinsic evaluation metrics suffer from over-sensitivity to their corresponding cosine similarity thresholds, indicating that a wider range of metrics need to be developed to track progress.
\end{abstract}

\section{Introduction}
Word embeddings have great practical importance since they can be used as pre-computed high-density \emph{features} to ML models, significantly reducing the amount of training data required in a variety of NLP tasks. However, there are several inter-related challenges with computing and consistently distributing word embeddings concerning the:
\begin{itemize}
\item \textbf{intrinsic properties of the embeddings}. How many dimensions do we actually need to store all the ``useful" semantic information? How big should the embedded vocabulary be to have practical value? How do these two factors interplay?
\item \textbf{type of model} used for generating the embeddings. There are multiple possible models and it is not obvious which one is the ``best", both in general or in the context of a specific type of applications.
\item the size and properties of \textbf{training data}: What is the minimum amount of training data needed? Should we include out of vocabulary words in the training?
\item optimization techniques to be used, \textbf{model hyperparameter} and \emph{training parameters}.
\end{itemize}
Not only the space of possibilities for each of these aspects is large, there are also challenges in performing a consistent large-scale evaluation of the resulting embeddings \cite{levy2015improving}. This makes systematic experimentation of alternative word-embedding configurations extremely difficult. 

In this work, we make progress in trying to find good combinations of some of the previous parameters. We focus specifically in the task of computing word embeddings for processing the Portuguese Twitter stream. User-generated content (such as twitter messages) tends to be populated by words that are specific to the medium, and that are constantly being added by users. These dynamics pose challenges to NLP systems, which have difficulties in dealing with out of vocabulary words. Therefore, learning a semantic representation for those words directly from the user-generated stream - and as the words arise - would allow us to keep up with the dynamics of the medium and reduce the cases for which we have no information about the words.

Starting from our own implementation of a neural word embedding model, which should be seen as a flexible baseline model for further experimentation, our research tries to answer the following practical questions:

\begin{itemize}
\item how large is the vocabulary the one can realistically embed given the level of resources that most organizations can afford to buy and to manage (as opposed to large clusters of GPU's only available to a few organizations)?
\item how much data, as a function of the size of the vocabulary we wish to embed, is enough for training meaningful embeddings?
\item how can we evaluate embeddings in automatic and consistent way so that a reasonably detailed systematic exploration of the previously describe space of possibilities can be performed?
\end{itemize}

By answering these questions based on a reasonably small sample of Twitter data (5M), we hope to find the best way to proceed and train embeddings for Twitter vocabulary using the much larger amount of Twitter data available (300M), but for which parameter experimentation would be unfeasible. This work can thus be seen as a \emph{preparatory study} for a subsequent attempt to produce and distribute a large-scale database of embeddings for processing Portuguese Twitter data.



\section{Related Work}

There are several approaches to generating word embeddings. One can build models that \emph{explicitly} aim at generating word embeddings, such as Word2Vec or GloVe \cite{mikolov2013distributed,pennington2014glove}, or one can extract such embeddings as by-products of more general models, which implicitly compute such word embeddings in the process of solving other language tasks.

Word embeddings methods aim to represent words as real valued continuous vectors in a much lower dimensional space when compared to traditional bag-of-words models. Moreover, this low dimensional space is able to capture lexical and semantic properties of words. Co-occurrence statistics are the fundamental information that allows creating such representations. Two approaches exist for building word embeddings. One creates a low rank approximation of the word co-occurrence matrix, such as in the case of Latent Semantic Analysis \cite{deerwester1990indexing} and GloVe \cite{pennington2014glove}. The other approach consists in extracting internal representations from neural network models of text \cite{bengio2003neural,collobert2008unified,mikolov2013distributed}. Levy and Goldberg \cite{levy2014neural} showed that the two approaches are closely related.  

Although, word embeddings research go back several decades, it was the recent developments of Deep Learning and the word2vec framework \cite{mikolov2013distributed} that captured the attention of the NLP community. Moreover, Mikolov et al. \cite{mikolov2013linguistic} showed that embeddings trained using word2vec models (CBOW and Skip-gram) exhibit linear structure, allowing analogy questions of the form ``man:woman::king:??.'' and can boost performance of several text classification tasks.

One of the issues of recent work in training word embeddings is the variability of experimental setups reported. For instance, in the paper describing GloVe \cite{pennington2014glove} authors trained their model on five corpora of different sizes and built a vocabulary of 400K most frequent words. Mikolov et al. \cite{mikolov2013linguistic} trained with 82K vocabulary while Mikolov et al. \cite{mikolov2013distributed} was trained with 3M vocabulary. Recently, Arora et al. \cite{arora2015rand} proposed a generative model for learning embeddings that tries to explain some theoretical justification for nonlinear models (e.g. word2vec and GloVe) and some hyper parameter choices. Authors evaluated their model using 68K vocabulary. 

SemEval 2016-Task 4: Sentiment Analysis in Twitter organizers report that participants either used general purpose pre-trained word embeddings, or trained from Tweet 2016 dataset or ``from some sort of dataset'' \cite{nakov2016semeval}. However, participants neither report the size of vocabulary used neither the possible effect it might have on the task specific results.

Recently, Rodrigues et al. \cite{rodrigues2016lx} created and distributed the first general purpose embeddings for Portuguese. Word2vec gensim implementation was used and authors report results with different values for the parameters of the framework. Furthermore, authors used experts to translate well established word embeddings test sets for Portuguese language, which they also made publicly available and we use some of those in this work.

\section{Our Neural Word Embedding Model}

The neural word embedding model we use in our experiments is heavily inspired in the one described in \cite{bengio2003neural}, but ours is one layer deeper and is set to solve a slightly different word prediction task. Given a sequence of 5 words - $w_{i-2}$ $w_{i-1}$ $w_i$ $w_{i+1}$ $w_{i+2}$, the task the model tries to perform is that of predicting the middle word, $w_i$, based on the two words on the left - $w_{i-2}$  $w_{i-1}$ - and the two words on the right - $w_{i+1}$ $w_{i+2}$: $P(w_i |w_{i-2}, w_{i-1}, w_{i+1}, w_{i+2})$. This should produce embeddings that closely capture distributional similarity, so that words that belong to the same semantic class, or which are synonyms and antonyms of each other, will be embedded in ``close'' regions of the embedding hyper-space.

Our neural model is composed of the following layers:
\begin{itemize}
\item a \textbf{Input Word Embedding Layer}, that maps each of the 4 input words represented by a 1-hot vectors with $|V|$ dimensions (e.g. 32k) into a low dimension space (64 bits). The projections matrix - $W_{input}$ - is shared across the 4 inputs. This is \emph{not} be the embedding matrix that we wish to produce.
\item a \textbf{Merge Layer} that \emph{concatenates} the 4 previous embeddings into a single vector holding all the context information. The concatenation operation ensures that the rest of the model has explicit information about the \textbf{relative position} of the input words. Using an \emph{additive} merge operation instead would preserve information onlu about the presence of the words, not their sequence.
\item a \textbf{Intermediate Context Embedding Dense Layer} that maps the preceding representation of 4 words into a lower dimension space, still representing the entire context. We have fixed this context representation to 64 dimensions. This ultimately determines the dimension of the resulting embeddings. This intermediate layer is important from the point of view of performance because it isolates the still relatively high-dimensional input space (4 x 64 bits input word embeddings) from the very high-dimensional output space.    
\item a final \textbf{Output Dense Layer} that maps the takes the previous 64-bit representation of the entire input context and produces a vector with the dimensionality of the word output space ($|V|$ dimensions). This matrix - $W_{output}$ - is the one that stores the word embeddings we are interested in.
\item A \textbf{Softmax Activation Layer} to produces the final prediction over the word space, that is the $P(w_i |w_{i-2}, w_{i-1}, w_{i+1}, w_{i+2})$ distribution
\end{itemize}
All neural activations in the model are sigmoid functions. The model was implemented using the Syntagma\footnote{https://github.com/sarmento/syntagma} library which relies on Keras~\cite{chollet2015} for model development, and we train the model using the built-in ADAM~\cite{kingma2014adam} optimizer with the default parameters. 

\section{Experimental Setup} \label{expsetup}
We are interested in assessing two aspects of the word embedding process. On one hand, we wish to evaluate the semantic quality of the produced embeddings. On the other, we want to quantify how much computational power and training data are required to train the embedding model as a function of the size of the vocabulary $|V|$ we try to embed. These aspects have fundamental practical importance for deciding how we should attempt to produce the large-scale database of embeddings we will provide in the future. All resources developed in this work are publicly available\footnote{https://github.com/saleiro/embedpt}.

Apart from the size of the vocabulary to be processed ($|V|$), the hyperparamaters of the model that we could potentially explore are i) the dimensionality of the input word embeddings and  ii) the dimensionality of the output word embeddings. As mentioned before, we set both to 64 bits after performing some quick manual experimentation. Full hyperparameter exploration is left for future work. 

Our experimental testbed comprises a desktop with a nvidia TITAN X (Pascal), Intel Core Quad i7 3770K 3.5Ghz, 32 GB DDR3 RAM and a 180GB SSD drive.

\subsection{Training Data}
We randomly sampled 5M tweets from a corpus of 300M tweets collected from the Portuguese Twitter community \cite{bovsnjak2012twitterecho}. The 5M comprise a total of 61.4M words (approx. 12 words per tweets in average). From those 5M tweets we generated a database containing 18.9M distinct 5-grams, along with their frequency counts. In this process, all text was down-cased. To help anonymizing the n-gram information, we substituted all the twitter handles by an artificial token ``T\_HANDLE". We also substituted all HTTP links by the token ``LINK". We prepended two special tokens to complete the 5-grams generated from the first two words of the tweet, and we correspondingly appended two other special tokens to complete 5-grams centered around the two last tokens of the tweet.

Tokenization was perform by \emph{trivially} separating tokens by blank space. No linguistic pre-processing, such as for example separating punctuation from words, was made. We opted for not doing any pre-processing for not introducing any linguistic bias from another tool (tokenization of user generated content is not a trivial problem). The most direct consequence of not performing any linguistic pre-processing is that of increasing the vocabulary size and diluting token counts. However, in principle, and given enough data, the embedding model should be able to learn the correct embeddings for both actual words (e.g. ``ronaldo") and the words that have punctuation attached (e.g. ``ronaldo!"). In practice, we believe that this can actually be an advantage for the downstream consumers of the embeddings, since they can also relax the requirements of their own tokenization stage. Overall, the dictionary thus produced contains approximately 1.3M distinct entries. Our dictionary was sorted by frequency, so the words with lowest index correspond to the most common words in the corpus.

We used the information from the 5-gram database to generate all training data used in the experiments. For a fixed size $|V|$ of the target vocabulary to be embedded (e.g. $|V|$ = 2048), we scanned the database to obtain \emph{all} possible 5-grams for which all tokens were among the top $|V|$ words of the dictionary (i.e. the top $|V|$ most frequent words in the corpus).  Depending on $|V|$, different numbers of valid training 5-grams were found in the database: the larger $|V|$ the more valid 5-grams would pass the filter. The number of examples collected for each of the values of $|V|$ is shown in Table \ref{size_training_data}. 
 
\begin{table}[t]
\centering
\caption{Number of 5-grams available for training for different sizes of target vocabulary $|V|$}
\label{size_training_data}
\begin{tabular}{|l|l|}
\hline
$|V|$    & \# 5-grams   \\ \hline
2048   & 2,496,830 \\ \hline
8192   & 6,114,640 \\ \hline
32768  & 10,899,570 \\ \hline
\end{tabular}
\end{table}

Since one of the goals of our experiments is to understand the impact of using different amounts of training data, for each size of vocabulary to be embedded $|V|$ we will run experiments training the models using 25\%, 50\%, 75\% and 100\% of the data available. 

\subsection{Metrics related with the Learning Process}

We tracked metrics related to the learning process itself, as a function of the vocabulary size to be embedded $|V|$ and of the fraction of training data used (25\%, 50\%, 75\% and 100\%). For all possible configurations, we recorded the values of the training and validation loss (cross entropy) after each epoch. Tracking these metrics serves as a minimalistic sanity check: if the model is not able to solve the word prediction task with some degree of success (e.g. if we observe no substantial decay in the losses) then one should not expect the embeddings to capture any of the distributional information they are supposed to capture.

\subsection{Tests and Gold-Standard Data for Intrinsic Evaluation}
Using the gold standard data (described below), we performed three types of tests:

\begin{itemize}
\item \textbf{Class Membership Tests}: embeddings corresponding two member of the same semantic class (e.g. ``Months of the Year", ``Portuguese Cities", ``Smileys") should be close, since they are supposed to be found in mostly the same contexts.
\item \textbf{Class Distinction Test}: this is the reciprocal of the previous Class Membership test. Embeddings of elements of different classes should be different, since words of different classes ere expected to be found in significantly different contexts.
\item \textbf{Word Equivalence Test}: embeddings corresponding to \emph{synonyms}, \emph{antonyms}, \emph{abbreviations} (e.g. ``porque" abbreviated by ``pq") and \emph{partial references} (e.g. ``slb and benfica") should be almost equal, since both alternatives are supposed to be used be interchangeable in all contexts (either maintaining or inverting the meaning).
\end{itemize}

Therefore, in our tests, two words are considered: 
\begin{itemize}
\item \emph{distinct} if the cosine of the corresponding embeddings is lower than \textbf{0.70} (or \textbf{0.80}). 
\item to \emph{belong to the same class} if the cosine of their embeddings is higher than \textbf{0.70} (or \textbf{0.80}).
\item equivalent if the cosine of the embeddings is higher that \textbf{0.85} (or \textbf{0.95}).
\end{itemize}
We report results using different thresholds of cosine similarity as we noticed that cosine similarity is skewed to higher values in the embedding space, as observed in related work \cite{dinu2014improving,faruqui2016problems}.
We used the following sources of data for testing Class Membership:
\begin{itemize}
\item AP+Battig data. This data was collected from the evaluation data provided by \cite{rodrigues2016lx}. These correspond to 29 semantic classes.
\item Twitter-Class - collected manually by the authors by checking top most frequent words in the dictionary and then expanding the classes. These include the following 6 sets (number of elements in brackets): smileys (13), months (12), countries (6), names (19), surnames (14) Portuguese cities (9).
\end{itemize}
For the Class Distinction test, we pair each element of each of the gold standard classes, with all the other elements from other classes (removing duplicate pairs since ordering does not matter), and we generate pairs of words which are supposed belong to different classes. For Word Equivalence test, we manually collected equivalente pairs, focusing on abbreviations that are popular in Twitters (e.g. ``qt" $\simeq$ ``quanto" or ``lx" $\simeq$ ``lisboa" and on frequent acronyms  (e.g. ``slb" $\simeq$ ``benfica"). In total, we compiled 48 equivalence pairs.

For all these tests we computed a \emph{coverage} metric. Our embeddings do not necessarily contain information for all the words contained in each of these tests. So, for all tests, we compute a \emph{coverage} metric that measures the fraction of the gold-standard pairs that could actually be tested using the different embeddings produced. Then, for all the test pairs actually covered, we obtain the success metrics for each of the 3 tests by computing the ratio of pairs we were able to correctly classified as i) being distinct (cosine $<$ 0.7 or 0.8), ii) belonging to the same class (cosine $>$ 0.7 or 0.8), and iii) being equivalent (cosine $>$ 0.85 or 0.95).

It is worth making a final comment about the gold standard data. Although we do not expect this gold standard data to be sufficient for a wide-spectrum evaluation of the resulting embeddings, it should be enough for providing us clues regarding areas where the embedding process is capturing enough semantics, and where it is not. These should still provide valuable indications for planning how to produce the much larger database of word embeddings.

\section{Results and Analysis}

We run the training process and performed the corresponding evaluation for 12 combinations of size of vocabulary to be embedded, and the volume of training data available that has been used. Table \ref{tab1} presents some overall statistics after training for 40 epochs.

\begin{table}[t]
\centering
\caption{Overall statistics for 12 combinations of models learned varying $|V|$ and volume of training data. Results observed after 40 training epochs.}
\label{tab1}
\begin{tabular}{|l|l|l|l|l|}
\hline
Embeddings                & \# Training Data Tuples & Avg secs/epoch & Training loss & Validation loss \\ \hline
$|V|$ = 2048  & 561,786  (25\% data)     & 4                  & 3.2564        & 3.5932          \\ \hline
$|V|$ = 2048  & 1,123,573 (50\% data)    & 9                  & 3.2234        & 3.4474          \\ \hline
$|V|$ = 2048    & 1,685,359 (75\% data)    & 13                 & 3.2138        & 3.3657          \\ \hline
$|V|$ = 2048   & 2,496,830  (100\% data)   & 18                 & 3.2075        & 3.3074          \\ \hline
$|V|$ = 8192    & 1,375,794 (25\% data)    & 63                 & 3.6329        & 4.286           \\ \hline
$|V|$ = 8192    & 2,751,588 (50\% data)     & 151                & 3.6917        & 4.0664          \\ \hline
$|V|$ = 8192    & 4,127,382 (75\% data)     & 187                & 3.7019        & 3.9323          \\ \hline
$|V|$ = 8192  & 6,114,640 (100\% data)    & 276                & 3.7072        & 3.8565          \\ \hline
$|V|$ = 32768   & 2,452,402 (25\% data)    & 388                & 3.7417        & 5.2768          \\ \hline
$|V|$ = 32768   & 4,904,806 (50\% data)    & 956                & 3.9885        & 4.8409          \\ \hline
$|V|$ = 32768   & 7,357,209 (75\% data)    & 1418               & 4.0649        & 4.6             \\ \hline
$|V|$ = 32768  & 10,899,570 (100\% data)    & 2028               & 4.107         & 4.4491          \\ \hline
\end{tabular}
\end{table}
\begin{figure}[t] \label{fig1}
  \begin{minipage}[b]{0.50\textwidth}
    \includegraphics[width=\textwidth]{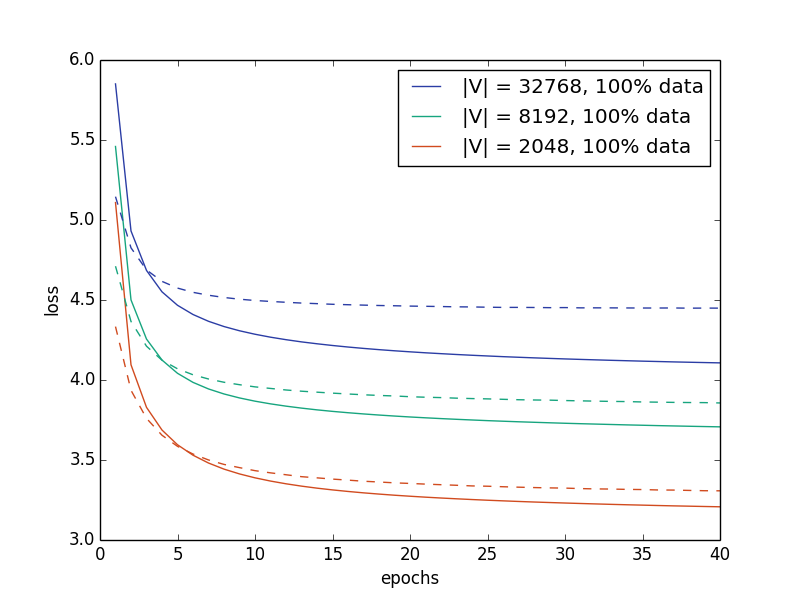}
  \end{minipage}
  \hfill
  \begin{minipage}[b]{0.50\textwidth}
    \includegraphics[width=\textwidth]{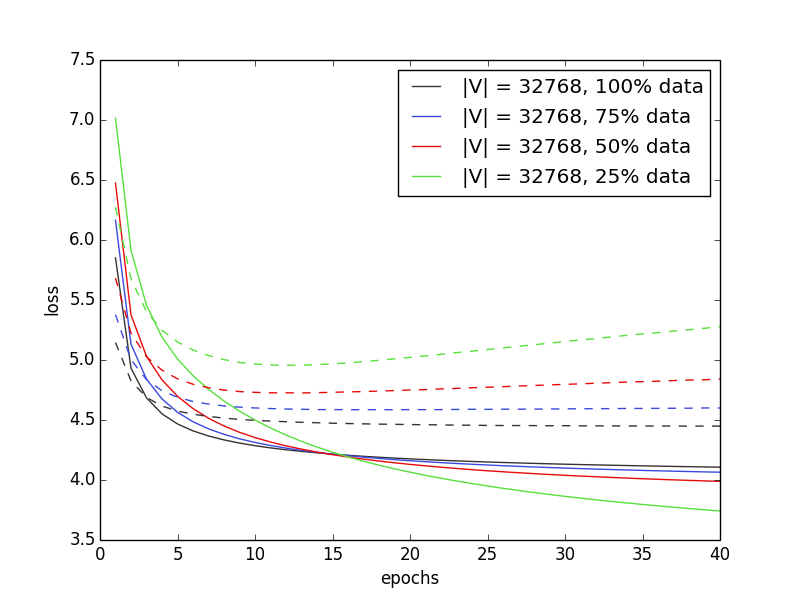}
  \end{minipage}
  \caption{Continuous line represents loss in the training data while dashed line represents loss in the validation data. Left side: effect of increasing $|V|$ using 100\% of training data. Right side: effect of varying the amount of training data used with $|V|$ = 32768.}
\end{figure} 
 The average time per epoch increases first with the size of the vocabulary to embed $|V|$ (because the model will have more parameters), and then, for each $|V|$, with the volume of training data. Using our testbed (Section \ref{expsetup}), the total time of learning in our experiments varied from a minimum of 160 seconds, with $|V|$ = 2048 and 25\% of data, to a maximum of 22.5 hours, with $|V|$ = 32768 and using 100\% of the training data available (extracted from 5M tweets). These numbers give us an approximate figure of how time consuming it would be to train embeddings from the complete Twitter corpus we have, consisting of 300M tweets.

We now analyze the learning process itself.  We plot the training set loss and validation set loss for the different values of $|V|$ (Figure \ref{fig1} left) with 40 epochs and using all the available data. As expected, the loss is reducing after each epoch, with validation loss, although being slightly higher, following the same trend. When using 100\% we see no model overfitting. We can also observe that the higher is $|V|$ the higher are the absolute values of the loss sets. This is not surprising because as the number of words to predict becomes higher the problem will tend to become harder. Also, because we keep the dimensionality of the embedding space constant (64 dimensions), it becomes increasingly hard to represent and differentiate larger vocabularies in the same hyper-volume. We believe this is a specially valuable indication for future experiments and for deciding the dimensionality of the final embeddings to distribute.

On the right side of Figure \ref{fig1} we show how the number of training (and validation) examples affects the loss. For a fixed $|V|$ = 32768 we varied the amount of data used for training from 25\% to 100\%. Three trends are apparent. As we train with more data, we obtain better validation losses. This was expected. The second trend is that by using less than 50\% of the data available the model tends to overfit the data, as indicated by the consistent increase in the validation loss after about 15 epochs (check dashed lines in right side of Figure \ref{fig1}). This suggests that for the future we should not try any drastic reduction of the training data to save training time. Finally, when not overfitting, the validation loss seems to stabilize after around 20 epochs. We observed no phase-transition effects (the model seems simple enough for not showing that type of behavior). This indicates we have a practical way of safely deciding when to stop training the model.

\subsection{Intrinsic Evaluation}
Table \ref{tabres} presents results for the three different tests described in Section \ref{expsetup}. The first (expected) result is that the coverage metrics increase with the size of the vocabulary being embedded, i.e., $|V|$. Because the Word Equivalence test set was specifically created for evaluating Twitter-based embedding, when embedding $|V|$ = 32768 words we achieve almost 90\% test coverage. On the other hand, for the Class Distinction test set - which was created by doing the cross product of the test cases of each class in Class Membership test set - we obtain very low coverage figures. This indicates that it is not always possible to re-use previously compiled gold-standard data, and that it will be important to compile gold-standard data directly from Twitter content if we want to perform a more precise evaluation. 

The effect of varying the cosine similarity decision threshold from 0.70 to 0.80 for Class Membership test shows that the percentage of classified as correct test cases drops significantly. However, the drop is more accentuated when training with only a portion of the available data. The differences of using two alternative thresholds values is even higher in the Word Equivalence test. 

The Word Equivalence test, in which we consider two words equivalent word if the cosine of the embedding vectors is higher than 0.95, revealed to be an extremely demanding test. Nevertheless, for $|V|$ = 32768 the results are far superior, and for a much larger coverage, than for lower $|V|$. The same happens with the Class Membership test. 

On the other hand, the Class Distinction test shows a different trend for larger values of $|V|$ = 32768 but the coverage for other values of $|V|$ is so low that becomes difficult to hypothesize about the reduced values of True Negatives (TN) percentage obtained for the largest $|V|$. It would be necessary to confirm this behavior with even larger values of $|V|$. One might hypothesize that the ability to distinguish between classes requires larger thresholds when $|V|$ is large. Also, we can speculate about the need of increasing the number of dimensions to be able to encapsulate different semantic information for so many words.

\begin{table}[t]
\caption{Evaluation of resulting embeddings using Class Membership, Class Distinction and Word Equivalence tests for different thresholds of cosine similarity.}
\label{tabres}
\centering
\begin{tabular}{|l|l|l|l|l|l|l|l|l|l|}
\hline
\multicolumn{1}{|c|}{Embeddings}      & \multicolumn{3}{c|}{Class Membership}                                                                                                                                                    & \multicolumn{3}{c|}{Class Distinction}                                                                                                                                            & \multicolumn{3}{c|}{Word Equivalence}                                                                                                                                                  \\ \hline
\multicolumn{1}{|c|}{$|V|$, \%data} & \multicolumn{1}{c|}{coverage} & \multicolumn{1}{c|}{\begin{tabular}[c]{@{}c@{}}Acc.\\ @0.70\end{tabular}} & \multicolumn{1}{c|}{\begin{tabular}[c]{@{}c@{}}Acc.\\  @0.80\end{tabular}} & \multicolumn{1}{c|}{coverage} & \multicolumn{1}{c|}{\begin{tabular}[c]{@{}c@{}}TN\\ @0.70\end{tabular}} & \multicolumn{1}{c|}{\begin{tabular}[c]{@{}c@{}}TN\\ @0.80\end{tabular}} & \multicolumn{1}{c|}{coverage} & \multicolumn{1}{c|}{\begin{tabular}[c]{@{}c@{}}Acc.\\ @0.85\end{tabular}} & \multicolumn{1}{c|}{\begin{tabular}[c]{@{}c@{}}Acc.\\ @0.95\end{tabular}} \\ \hline
2048, 25\%                            & \multirow{4}{*}{12.32\%}      & 30.71\%                                                                    & 4.94\%                                                                      & \multirow{4}{*}{1.20\%}       & 100\%                                                                   & 100\%                                                                   & \multirow{4}{*}{31.25\%}      & 26.67\%                                                                    & 2.94\%                                                                     \\ \cline{1-1} \cline{3-4} \cline{6-7} \cline{9-10} 
2048, 50\%                            &                               & 29.13\%                                                                    & 12.69\%                                                                     &                               & 100\%                                                                   & 100\%                                                                   &                               & 26.67\%                                                                    & 2.94\%                                                                     \\ \cline{1-1} \cline{3-4} \cline{6-7} \cline{9-10} 
2048, 75\%                            &                               & 29.13\%                                                                    & 18.12\%                                                                     &                               & 100\%                                                                   & 100\%                                                                   &                               & 33.33\%                                                                    & 2.94\%                                                                     \\ \cline{1-1} \cline{3-4} \cline{6-7} \cline{9-10} 
2048, 100\%                           &                               & 32.28\%                                                                    & 26.77\%                                                                     &                               & 100\%                                                                   & 100\%                                                                   &                               & 33.33\%                                                                    & 6.67\%                                                                     \\ \hline
8192, 25\%                            & \multirow{4}{*}{29.60\%}      & 14.17\%                                                                    & 4.94\%                                                                      & \multirow{4}{*}{6.54\%}       & 100\%                                                                   & 100\%                                                                   & \multirow{4}{*}{70.83\%}      & 14.71\%                                                                    & 2.94\%                                                                     \\ \cline{1-1} \cline{3-4} \cline{6-7} \cline{9-10} 
8192, 50\%                            &                               & 22.41\%                                                                    & 12.69\%                                                                     &                               & 99\%                                                                    & 100\%                                                                   &                               & 20.59\%                                                                    & 2.94\%                                                                     \\ \cline{1-1} \cline{3-4} \cline{6-7} \cline{9-10} 
8192, 75\%                            &                               & 27.51\%                                                                    & 18.12\%                                                                     &                               & 99\%                                                                    & 100\%                                                                   &                               & 20.59\%                                                                    & 2.94\%                                                                     \\ \cline{1-1} \cline{3-4} \cline{6-7} \cline{9-10} 
8192, 100\%                           &                               & 33.77\%                                                                    & 21.91\%                                                                     &                               & 97\%                                                                    & 100\%                                                                   &                               & 29.41\%                                                                    & 5.88\%                                                                     \\ \hline
32768, 25\%                           & \multirow{4}{*}{47.79\%}      & 17.73\%                                                                    & 5.13\%                                                                      & \multirow{4}{*}{18.31\%}      & 98\%                                                                    & 100\%                                                                   & \multirow{4}{*}{89.58\%}      & 16.28\%                                                                    & 2.33\%                                                                     \\ \cline{1-1} \cline{3-4} \cline{6-7} \cline{9-10} 
32768, 50\%                           &                               & 52.30\%                                                                    & 21.06\%                                                                     &                               & 83\%                                                                    & 98\%                                                                    &                               & 34.88\%                                                                    & 9.30\%                                                                     \\ \cline{1-1} \cline{3-4} \cline{6-7} \cline{9-10} 
32768, 75\%                           &                               & 85.15\%                                                                    & 49.41\%                                                                     &                               & 44\%                                                                    & 88\%                                                                    &                               & 58.14\%                                                                    & 23.26\%                                                                    \\ \cline{1-1} \cline{3-4} \cline{6-7} \cline{9-10} 
32768, 100\%                          &                               & 95.59\%                                                                    & 74.80\%                                                                     &                               & 13\%                                                                    & 57\%                                                                    &                               & 72.09\%                                                                    & 34.88\%                                                                    \\ \hline
\end{tabular}

\end{table}

\subsection{Further Analysis regarding Evaluation Metrics}

Despite already providing interesting practical clues for our goal of trying to embed a larger vocabulary using more of the training data we have available, these results also revealed that the intrinsic evaluation metrics we are using are overly sensitive to their corresponding cosine similarity thresholds. This sensitivity poses serious challenges for further systematic exploration of word embedding architectures and their corresponding hyper-parameters, which was also observed in other recent works \cite{faruqui2016problems}. 

By using these absolute thresholds as criteria for deciding similarity of words, we create a dependency between the evaluation metrics and the \emph{geometry} of the embedded data. If we see the embedding data as a graph, this means that metrics will change if we apply scaling operations to certain parts of the graph, even if its structure (i.e. relative position of the embedded words) does not change. 

For most practical purposes (including training downstream ML models) absolute distances have little meaning. What is fundamental is that the resulting embeddings are able to capture topological information: similar words should be \emph{closer to each other} than they are to words that are dissimilar to them (under the various criteria of similarity we care about), independently of the absolute distances involved.

It is now clear that a key aspect for future work will be developing additional performance metrics based on topological properties. We are in line with recent work \cite{gladkova2016intrinsic}, proposing to shift evaluation from absolute values to more exploratory evaluations focusing on weaknesses and strengths of the embeddings and not so much in generic scores. For example, one metric could consist in checking whether for any given word, all words that are known to belong to the same class are closer than any words belonging to different classes, independently of the actual cosine. Future work will necessarily include developing this type of metrics.

\section{Conclusions}

Producing word embeddings from tweets is challenging due to the specificities of the vocabulary in the medium. We implemented a neural word embedding model that embeds words based on n-gram information extracted from a sample of the Portuguese Twitter stream, and which can be seen as a flexible baseline for further experiments in the field. Work reported in this paper is a preliminary study of trying to find parameters for training word embeddings from Twitter and adequate evaluation tests and gold-standard data.

Results show that using less than 50\% of the available training examples for each vocabulary size might result in overfitting. The resulting embeddings obtain an interesting performance on intrinsic evaluation tests when trained a vocabulary containing the 32768 most frequent words in a Twitter sample of relatively small size. Nevertheless, results exhibit a skewness in the cosine similarity scores that should be further explored in future work. More specifically, the Class Distinction test set revealed to be challenging and opens the door to evaluation of not only similarity between words but also dissimilarities between words of different semantic classes without using absolute score values. 

Therefore, a key area of future exploration has to do with better evaluation resources and metrics. We made some initial effort in this front. However, we believe that developing new intrinsic tests, agnostic to absolute values of metrics and concerned with topological aspects of the embedding space, and expanding gold-standard data with cases tailored for user-generated content, is of fundamental importance for the progress of this line of work.

Furthermore, we plan to make public available word embeddings trained from a large sample of 300M tweets collected from the Portuguese Twitter stream. This will require experimenting producing embeddings with higher dimensionality (to avoid the cosine skewness effect) and training with even larger vocabularies. Also, there is room for experimenting with some of the hyper-parameters of the model itself (e.g. activation functions, dimensions of the layers), which we know have impact on final results. 

\subsubsection{Acknowledgements}
We gratefully acknowledge the support of NVIDIA Corporation with the donation of the Titan X Pascal GPU used for this research.

\bibliographystyle{unsrt}
\bibliography{refs}  

\end{document}